\documentclass[journal]{IEEEtran} 
\usepackage{times}
\usepackage{epsfig}
\usepackage{graphics}
\usepackage{amsmath}
\usepackage{float}
\usepackage{amssymb}
\usepackage{subfigure}
\usepackage{footmisc}
\usepackage{indentfirst}
\usepackage{fancyhdr}
\usepackage{multicol}
\usepackage{color}
\usepackage{fullpage,authblk}
\begin{document}
\title{\textbf{CSIFT Based Locality-constrained Linear Coding for Image Classification}}
\author{CHEN~Junzhou $^1$ 
, LI Qing $^1$, PENG Qiang $^1$ and Kin Hong Wong $^2$\\
\small{} $^1$ School of Information Science \& Technology, Southwest Jiaotong University, Chengdu, Sichuan, 610031, China\\
$^2$ Department of Computer Science \&  Engineering, The Chinese University of Hong Kong,  Shatin, Hong Kong\\
Email:jzchen@swjtu.edu.cn; khwong@cse.cuhk.edu.hk
}
\date{}
\maketitle

\centerline{\textbf{Abstract}}
In the past decade, SIFT descriptor has been witnessed as one of the most robust local invariant feature descriptors and widely used in various vision tasks. Most traditional image classification systems depend on the luminance-based SIFT descriptors, which only analyze the gray level variations of the images. Misclassification may happen since their color contents are ignored. In this article, we concentrate on improving the performance of existing image classification algorithms by adding color information. To achieve this purpose, different kinds of colored SIFT descriptors are introduced and implemented. \emph{Locality-constrained Linear Coding} (LLC), a state-of-the-art sparse coding technology, is employed to construct the image classification system for the evaluation. The real experiments are carried out on several benchmarks. With the enhancements of color SIFT, the proposed image classification system obtains approximate $3\%$ improvement of classification accuracy on the Caltech-101 dataset and approximate $4\%$ improvement of classification accuracy on the Caltech-256 dataset.
\section{Introduction}\label{sec:introduction}
S\emph{cale invariant feature transform} (SIFT) descriptors \cite{lowe2004distinctive} are widely used in many vision tasks, such as object recognition, image classification, video retrieval, etc. It has been witnessed a very robust local invariant feature descriptors in respect of different geometrical changes. However, SIFT was mainly developed for gray images, the color information of the objects are neglected. Therefore, two objects with completely different colors may be regarded as the same. To overcome this limitation, different kinds of \emph{Colored SIFT} (CSIFT) descriptors were proposed and developed by researchers to utilize the color information inside the SIFT descriptors \cite{geusebroek2001color} \cite{abdel2006csift} \cite{van2006boosting} \cite{burghouts2009performance} \cite{gevers2012color}. With the enhancement of color information, CSIFT descriptors can achieve better performances in resisting some certain photometric changes. One example can be found in \cite{abdel2006csift}, which shows that CSIFT is more stable than SIFT in case of illumination changes.

On the other hand, the \emph{bag-of-features} (BoF) \cite{goldfarb1983numerically} \cite{csurka2004visual} joined with the \emph{spatial pyramid matching} (SPM) kernel \cite{lazebnik2006beyond} has been employed to build the recent state-of-the-art image classification systems. In BoF, images are considered as sets of unordered local appearance descriptors, which are clustered into discrete “visual words” for the representation of images in semantic classification.

SPM divides an image into $2^l\times 2^l$ segments in different scales $l = 0, 1, 2$, computes the BoF histogram within each segment, and finally concatenates all the histograms to build a spatial location sensitive descriptor of the image. In order to obtain better classification performance, a codebook (a set of visual words), also named dictionary, is constructed to represent the extracted descriptors. Traditional SPM uses clustering techniques like K-means \emph{vector quantization} (VQ) to generate the codebook. Despite their efficiency, the obtained codebooks usually suffer from several drawbacks such as distortion errors and low discriminative ability \cite{shabou2012locality}. A linear \emph{SPM based on sparse coding} (ScSPM) method \cite{yang2009linear} was proposed by Yang \emph{et al}. to relaxing the restrictive cardinality constraint of VQ. By generalizing vector quantization to sparse coding followed by multi-scale spatial max-pooling, ScSPM significantly outperforms the traditional SPM kernel on histograms and is even better than the nonlinear SPM kernels on several benchmarks.

Yu \emph{et al}. \cite{yu2009nonlinear} demonstrated that under certain assumptions locality is more essential than sparsity for the training of nonlinear classifiers and proposed a modification of SC, named \emph{Local Coordinate Coding} (LCC). However, in both SC and LCC, the computationally expensive L1-norm optimization problem is to be solved. Wang \emph{et al}. developed a faster implementation of LCC, named \emph{locality-constrained linear coding} (LLC) \cite{wang2010locality}, which utilizes the locality constraint to project each descriptor into its local-coordinate system. It achieves the state-of-the-art image classification accuracy even just using a linear SVM classifier.

According to our literature survey, although various kinds of \emph{sparse representation} (SR) based image classification algorithms with state-of-the-art performances have been developed, most of them use only luminance-based SIFT descriptors \cite{yang2009linear} \cite{wang2010locality} \cite{yang2010supervised} \cite{liu2011defense} \cite{yang2011fisher} \cite{shabou2012locality}. Using color information can improve the robustness of traditional SIFT descriptor in respect of color variations and the geometrical changes. However, facing the diverse CSIFT descriptors, the following questions are worth to be studied.
\begin{itemize}
  \item Which CSIFT descriptor is the best for the SR based image classification system?
  \item In what extend, the performance of SR based image classification system can be improved by using CSIFT?
\end{itemize}

To fully exploit the potential of CSIFT descriptors for image category recognition tasks, a CSIFT based image classification system is constructed in this work. As a widely used state-of-the-art SC based encoding algorithm, LLC is employed to encode the CSIFT descriptors for classification. Real experiments with different kinds of CSIFT descriptors demonstrate that significant improvements can be obtained with the enhancement of color information.

The rest of this article is organized as follows. In section \ref{sec:the reflectance model}, a reflectance model for color analysis is presented. In section \ref{sec:Descriptors}, different kinds of the CSIFT descriptors and their properties are discussed. Section \ref{sec:LLC} introduces the basic concepts of the LLC. In section \ref{sec:Exp} and \ref{sec:Further}, real experiments are carried out to study the proposed algorithm in various aspects. Finally, in
section \ref{sec:conclusion}, conclusions are drawn. 
\section{Dichromatic Reflectance Model} \label{sec:the reflectance model}
A physical model of reflection, named \emph{Dichromatic Reflection Model}, was presented by Shafer in 1985 \cite{shafer1985using}. In which, the relationship between RGB-values of captured images and the photometric changes, such as shadows and specularities, of environment was investigated. Shafer indicated that the reflection of a incident light can be divided into two distinct components: specular reflection and body reflection. Specular reflection is when a ray of light hits a smooth surface at certain angle. The reflection of that ray will reflect at the same angle as the incident ray. The effect of highlight is caused by the specular reflection. Diffuse reflection is when a ray of light hits the surface which will be reflected back in every direction.

Consider an image of an infinitesimal surface patch of some object. Let the red, green and blue sensors with spectral sensitivities be $f_{R}(\lambda)$, $f_{G}(\lambda)$ and $f_{B}(\lambda)$ respectively. The corresponding sensor values of the surface image are \cite{shafer1985using} \cite{gevers2007color}:
\begin{equation}\label{eq:Lmodel1}
\begin{split}
    \begin{array}{rcl}
      L(\lambda,\textbf{n},\textbf{s},\textbf{v})&=&m_{b}(\textbf{n},\textbf{s})\int_{\lambda}{f_{L}(\lambda)e(\lambda)c_{b}(\lambda)\, d\lambda}+ \\
                      && m_{s}(\textbf{n},\textbf{s},\textbf{v})\int_{\lambda}{f_{L}(\lambda)e(\lambda)c_{s}(\lambda)\, d\lambda}
    \end{array}
\end{split}
\end{equation}
where $L\in \{R, G,B\}$ is the color channel of light, $\lambda$ is the wavelength, $\textbf{n}$ is the surface patch normal, $\textbf{s}$ is the direction of the illumination source, and $\textbf{v}$ is the direction of the viewer. $e(\lambda)$ is power of the incident light with wavelength $\lambda$, $c_{b}(\lambda)$ and $c_{s}$  are the the surface albedo and Fresnel reflectance, respectively. The geometric terms $m_b$ and $m_s$ represent the diffuse reflection and the specular reflection respectively.

In case white illumination and neutral interface reflection model holds, the incident light energy $e(\lambda) = e$ and Fresnel reflectance term $c_s(\lambda) = c_s$ are both constant values independent of the wavelength $\lambda$. By assuming the following holds:
\begin{equation}\label{eq:f_constant}
    \int_{\lambda}{f_{R}(\lambda)} = \int_{\lambda}{f_{G}(\lambda)}= \int_{\lambda}{f_{B}(\lambda)} = f
\end{equation}

Eq. (\ref{eq:Lmodel1}) can be simplified:

\begin{equation}\label{eq:Lmodel2}
      L(\textbf{n},\textbf{s},\textbf{v})=e m_{b}(\textbf{n},\textbf{s})k_L+ e m_{s}(\textbf{n},\textbf{s},\textbf{v})c_{s}f
\end{equation}
where $k_L = \int_{\lambda}f_L(\lambda)c_b(\lambda)$ is a variable depends only on the sensors and the surface albedo. 
\section{Colored SIFT Descriptors} \label{sec:Descriptors}
On the basis of the \emph{Dichromatic Reflection Model}, the stabilities and reliabilities  of color spaces in regard of various photometric events such as shadows and specularities are studied in both theoretically and empirically \cite{gevers1999color} \cite{geusebroek2001color} \cite{van2010evaluating}.
Although there are many existing color space models, they are correlated to intensity; are linear combinations of $RGB$; or normalized with respect to intensity rgb \cite{gevers1999color}. In this article, we concentrate on investigating CSIFT using essentially different color spaces: \emph{RGB}, \emph{HSV},  \emph{YCbCr}, \emph{Opponent}, \emph{rg} and color invariant spaces.
\subsection{SIFT} \label{sec:A}
The SIFT algorithm was originally developed for grey images by Lowe  \cite{lowe1999object} \cite{lowe2004distinctive} for extracting highly discriminative local
image features that are invariant to image scaling and rotation, and partially invariant to changes in illumination and viewpoint. It has been used in a broad range of vision tasks, such as image classification, recognition, content-based image-retrieval, etc. The algorithm involves two steps: 1) extraction of
the keypoints of an image; 2) computation of the feature vectors characterizing the keypoints. The first step is carried out by convolving the input image with the DoG (difference of Gaussians) function in multiple scales and detecting the extremas of the outputs. The second step is achieved by sampling the magnitudes and orientations of the image gradient in a patch around the detected feature. A 128-D vector of direction histograms is finally constructed as the descriptor of each patch. Since the SIFT descriptor is normalized, it can invariant to the scale of gradient magnitude. But the light color changes will affect it, because the intensity channel is a combination of the R, G and B channels.
\subsection{RGB-SIFT} \label{sec:B}
As the most popular color model, \emph{RGB} color space provides plenty information for vision applications. In order to embed \emph{RGB} color information into the SIFT descriptor, we simply calculate the traditional SIFT descriptors on the each channel of \emph{RGB} color space. By combining the extracted feature, a $128\times3$ dimensions descriptor is built ($128$ for each color channel). Compared with conventional luminance-based SIFT, the \emph{RGB} color gradients (or edges) of the image are captured.
\subsection{HSV-SIFT} \label{sec:C}
HSV-SIFT was introduced by Bosch \emph{et al.} and employed for scene classification task \cite{bosch2008scene}. Similar to \emph{RGB} SIFT discussed above, they compute SIFT descriptors over all three channels of the HSV color model and produces a $128\times3$ dimensional SIFT descriptor for each point. It is worth mention that, H channel of \emph{HSV} color model has scale-invariant and shift-invariant with respect to light intensity. However, due to the combination of the HSV channels, the whole descriptor has no invariance properties. The conversion from RGB space to HSV space is defined by Eq. (\ref{eq:rgb2hsv1})(\ref{eq:rgb2hsv2})(\ref{eq:rgb2hsv3}).\\

\begin{equation}\label{eq:rgb2hsv1}
H = \left\{\begin{array}{@{}c p{2.5cm}}
            undefined & \textrm{if $max=min$}\\
            60^\circ\times\frac{G-B}{max-min}+0^\circ & \textrm{if $max=R$ and $G\geq B$}\\
            60^\circ\times\frac{G-B}{max-min}+360^\circ & \textrm{if $max=R$ and $G<B$}\\
            60^\circ\times\frac{G-B}{max-min}+120^\circ & \textrm{if $max=G$}\\
            60^\circ\times\frac{G-B}{max-min}+240^\circ & \textrm{if $max=B$}
           \end{array}
    \right.
\end{equation}

\begin{equation}\label{eq:rgb2hsv2}
S = \left\{\begin{array}{cl}
            0 & \textrm{if $max=0$}\\
            \frac{max-min}{max}=1-\frac{min}{max} & \textrm{$otherwise$}
           \end{array}
    \right.
\end{equation}
\begin{equation}\label{eq:rgb2hsv3}
    V=max
\end{equation}

where, $max$ is equal to the maximal one of $R,G,B$, and $min$ is equal to the minimal one of $R,G,B$.
\subsection{rg-SIFT} \label{sec:D}
The rg-SIFT descriptors are obtained from the \emph{rg} color space. It is the normalized RGB color model, used \emph{r} and \emph{g} channels to describe the color information in the image (\emph{b} is constant if \emph{r} and \emph{g} are given). \emph{rg} color space is already scale-invariant with respect to light intensity.  The conversion from RGB space to rg space is defined as follows,

\begin{equation}\label{eq:RGB2r}
r = \frac{R}{R+G+B}
\end{equation}
\begin{equation}\label{eq:RGB2g}
g = \frac{G}{R+G+B}
\end{equation}
\subsection{YCrCb-SIFT}\label{sec:E}
As one of the most popular color spaces, YCrCb color space provides very efficient representation of scenes / images and is widely used in the field of video compression. It represents colors in terms of one luminance component ($Y$), and two chrominance components ($C_b$ and $C_r$). The YCbCr-SIFT descriptors are computed on all the channels of YCbCr color space. The YCbCr image can be converted from RGB images using equation below:

\begin{equation}\label{eq:rgb2YCbCr}
\begin{bmatrix}
  Y \\
  Cb \\
  Cr
\end{bmatrix}
=
\begin{bmatrix}
  \frac{R-G}{\sqrt{2}} \\
   \frac{R+G-2B}{\sqrt{6}} \\
    \frac{R+G+B}{\sqrt{3}}
\end{bmatrix}
\end{equation}
\subsection{Opponent-SIFT}\label{sec:F}
The Opponent color space was first proposed by Ewald Hering in the late 19th century \cite{hering1964outlines}. It consists three channels ($O_1$, $O_2$, $O_3$), in which the $O_3$ channel represents luminance of the image, while the remainder describe the opponent color (red-green, blue-yellow) of the image. Opponent-SIFT descriptor is obtained by computing the SIFT descriptor over each channel of the Opponent color space and combine them together. The RGB images transform in the opponent color space is defined by Eq. (\ref{eq:rgb2Opponent}).\\

\begin{equation}\label{eq:rgb2Opponent}
\left[
  \begin{array}{@{}c@{}}
  o_1 \\
  o_2 \\
  o_3
  \end{array}
\right]
=
\left[
  \begin{array}{@{}ccc@{}}
   0.299 & 0.587 & 0.144 \\
   -0.1687 & -0.3313 & 0.5  \\
   0.5 & -0.4187 & -0.0813
  \end{array}
\right]
\left[
  \begin{array}{@{}c@{}}
  R \\
  G \\
  B
  \end{array}
\right]
+
\left[
  \begin{array}{@{}c@{}}
  0 \\
  128 \\
  128
  \end{array}
\right]
\end{equation}
\subsection{Color Invariant SIFT} \label{sec:E}
With the inspiration of Dichromatic Reflectance Model (see section \ref{sec:the reflectance model}), the color-based photometric invariant scheme was proposed by M. Geusebroek \cite{geusebroek2001color}. It was first applied to SIFT descriptor by Abdel-Hakim and Farag \cite{abdel2006csift}. A linear transformation from RGB to color invariant space is presented as the following:
\begin{equation}\label{eq:GOCmodel}
\begin{bmatrix}
    \hat{E}(x,y)\\
    \hat{E}_{\lambda}(x,y) \\
   \hat{E}_{\lambda\lambda}(x,y)
\end{bmatrix}
=
\begin{pmatrix}
0.06 & 0.63 & 0.27 \\
0.30 & 0.04 & 0.35  \\
0.34 & 0.60 & 0.17
\end{pmatrix}
\begin{bmatrix}
    R(x,y)\\
    G(x,y) \\
    B(x,y)
\end{bmatrix}
\end{equation}
Where $\hat{E}(x,y)$, $\hat{E}_{\lambda}(x,y)$, $\hat{E}_{\lambda\lambda}(x,y)$, denoting ,respectively, the intensity, the yellow-blue channel, and the red-green channel. $\hat{E}$, $\hat{E}_{\lambda}$ and $\hat{E}_{\lambda\lambda}$ are the spectral differential quotients, and represent as the same as the above. Measurement of the color invariants is obtained by $\hat{E}$, $\hat{E}_{\lambda}$ and $\hat{E}_{\lambda\lambda}$.

\section{Locality-constrained Linear Coding} \label{sec:LLC}
The \emph{bag-of-feature} (BoF) approach has now played a leading role in the field of generic image classification research \cite{yang2009linear} \cite{wang2010locality} \cite{liu2011defense}. It commonly consists of feature extraction, codebook construction, feature coding, and feature pooling. Previous experimentally results shown that, given a visual codebook, choosing an appropriate coding scheme has significant impact on the classification performance.

Different kinds of coding algorithms are developed \cite{yang2009linear} \cite{wang2010locality} \cite{liu2011defense} \cite{shabou2012locality}, among them, \emph{Locality-constrained Linear Coding} (LLC) \cite{wang2010locality} is considered as one of the most representative methods, which provides both fast coding speed and state-of-the-art classification accuracy. It has been widely cited in academic papers and employed in image classification applications. In this article, LLC is selected for feature coding in our real experiments.

Let $X$ denotes a set of $D$-dimensional local descriptors in an image, i.e. $X = [x_{1}, x_{2},\ldots,x_{N}]\in R^{D\times N}$. Let $B = [b_{1},b_{2},\ldots,b_{M}]\in R^{D\times M}$ be a visual codebook with $M$ entries. The coding methods convert each descriptor into a $M$-dimensional code. Unlike the sparse coding, LLC enforces locality constraint instead of sparse constraint. A reconstruction for the basis descriptors $B$ can be acquired by optimizing the following equation:
\begin{equation}\label{eq:LLCmodel1}
    \min_{v}\sum_{i=1}^{N}\|x_{i}-Bv_{i}\|^{2}+\lambda\|d_{i}\odot v_{i}\|^{2} \\
    \\
    s.t. 1^{T}v_{i}=1,\forall i
\end{equation}
where $\odot $ denotes the element-wise multiplication, and $d_{i}\in R^{M}$ is the locality adaptor that gives some degree of freedom for each basis descriptor. LLC ensures these descriptors are proportionally similar to the input descriptor $x_{i}$. Specifically,
\begin{equation}\label{eq:LLCmodel2}
    d_{i}=\exp[\frac{dist(x_{i},B)}{\sigma}]
\end{equation}
where $dist(x_{i},B)=[dist(x_{1},b_{1}),dist(x_{2},b_{2}),\ldots,$ $dist(x_{i},b_{M})]$, and $dist(x_{i},b_{j})$ is the Euclidean distance between $x_{i}$ and $b_{j}$.$\sigma$ is used for adjusting the weight decay speed for the locality adaptor $d_{i}$.

An approximation is proposed in \cite{wang2010locality} to accelerate its computational efficiency in practice by ignoring the second term in Eq.(\ref{eq:LLCmodel1}). They directly use the $K$ nearest basis descriptors of $x_{i}$ to minimize the first term. The encoding process is simplified by solving a much smaller linear system,
\begin{equation}\label{eq:LLCmodel3}
    \min_{v}\sum_{i=1}^{N}\|x_{i}-Bv_{i}\|^{2} \\
    \\
    s.t. 1^{T}v_{i}=1,\forall i
\end{equation}

This gives the coding coefficients for the selected k basis vectors and other coefficient are set to zero.
\section{Experimental Results} \label{sec:Exp}
To evaluate the performances of different kinds of the CSIFT descriptors in a \emph{sparse representation} based image classification system, two benchmark datasets: Caltech-101\cite{fei2006one} and Caltech-256 \cite{griffin2007caltech} are employed in the real experiment. Since color information is the prerequisite for the CSIFT descriptors computation, to achieve a fair comparison, the gray images in the  Caltech-101 and Caltech-256 are removed. To enable colored images of some categories are sufficient for training a stable classifier (the number of colored images less than 31), we add some new color images of the same category that is to make sure there are at least 31 colored images in each category.
\subsection{Implementation}
In all the experiments, the same processing chain similar the settings refereed in this literature is used to ensure consistency.
\begin{enumerate}
  \item \emph{Colored SIFT} (CSIFT) / SIFT descriptors extraction. The dense CSIFT/SIFT descriptors are extracted as described in section \ref{sec:Descriptors} within a regular spatial grid. The step-size is fixed to 8 pixels and patch size to $16 \times 16$ pixels. The dimension of luminance-based SIFT descriptor is $128$. For CSIFT descriptors, RGB-SIFT, SIFT, HSV-SIFT, YCbCr-SIFT, Opponent-SIFT, rg-SIFT and Color Invariance SIFT (C-SIFT) are implemented for the experimentation. 
  \item Codebooks construction. After the CSIFT/SIFT descriptors are extracted, a codebook of size 1024 is created using the K-means clustering method on a randomly selected subset (with size $2 \times 10^6$) of extracted CSIFT descriptors;
  \item \emph{Locality-constrained linear coding} (LLC). The CSIFT/SIFT descriptors are encoded by LLC using the above constructed coodbooks.  the number of neighbors is set to 5 with the shift-invariant constraint;
  \item Pooling with \emph{spatial pyramid matching} (SPM) \cite{lazebnik2006beyond}. The max-pooling operation is adopted to compute the final descriptor of each image. It is performed with a $3$ levels SPM kernel ($1 \times 1$, $2 \times 2$ and $4 \times 4$ sub-regions in the corresponding levels), leaving a same weight at each layer. The pooled features of the sub-regions are concatenated and normalized to form the final descriptor of each image;
  \item Classification. a one-vs-all linear SVM classifier \cite{fan2008liblinear} is used to train the classifier, since it has shown good performances. 
\end{enumerate}

\subsection{Assessment of Color Descriptors on the Caltech-101 Dataset} \label{sec:Caltech101}
\begin{figure}
  \includegraphics[width=3in]{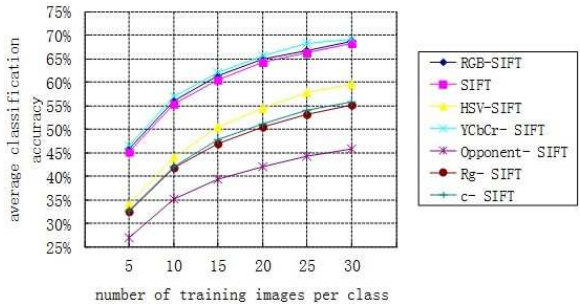}\\
  \caption{The different number of training images per class on the classification performance.}\label{fig:figure1}
\end{figure}

The propose algorithm is carried out using the color images of Caltech-101 dataset, which contains 101 object categories including animals, flowers, vehicles, shapes with significant variance, etc. Some color images are added to avoid insufficient of training data in certain categories as discussed before. The number of original images in every category still varies from 31 to 800. In order to test the performance with different sizes of training data, different numbers (5, 10, $\dots$, 30) of training images per category is evaluated. In each experiment, we randomly select $n$ images per category for training and leaving the remainders for testing. The images were resized to keep the maximum size of height and width no larger than 300 pixels with a conserved aspect ratio. For the sake of simplicity, the codebook size is fixed to 1024 (the performance of different codebook sizes will be studied in section \ref{sec:Codebook_size}). The corresponding results using different kinds of CSIFT descriptors (RGB-SIFT, SIFT, HSV-SIFT, YCbCr-SIFT, Opponent-SIFT, rg-SIFT and Color Invariance SIFT (C-SIFT)) are illustrated in Table \ref{tab:table1} and Figure \ref{fig:figure1}. According to the experimental results, all the CSIFT/SIFT descriptors achieve their best classification accuracy with 30 training images per class. It indicates that more training data may bring better classification accuracy in testing, while the improvement became slight when the size of the number of training images is more than $20$. Both RGB-SIFT and YCbCr-SIFT outperform state-of-the-art luminance-based SIFT on this dataset. The YCbCr-SIFT achieves the best performance. For instance, when 30 images of each category are used for training, YCbCr-SIFT obtains the average classification accuracy of $69.1\%$; RGB-SIFT provides the second best average classification accuracy ($68.6\%$). It is worth mentioning that even without color information, SIFT achieves third best average classification accuracy of $68.17\%$. Approximately $1\%$ improvement in average classification accuracy can be obtained by employing CSIFT descriptors.\\
\begin{table*}
\centering
\caption{Classification rate$(\%)$ comparison on Caltech-101}
\label{tab:table1}
\begin{tabular}{|l|c|c|c|c|c|c|} 
\hline \scriptsize Training images & \scriptsize 5 & \scriptsize 10 &\scriptsize 15 &\scriptsize 20 &\scriptsize 25 &\scriptsize 30 \\
\hline \scriptsize RGB-SIFT &\scriptsize 45.77 $\pm$ 1.02 & \scriptsize 55.90 $\pm$ 0.69 &\scriptsize 61.26$\pm$0.84 &\scriptsize 64.84$\pm$0.68 &\scriptsize 66.70$\pm$0.81 &\scriptsize 68.65$\pm$1.13 \\
\hline \scriptsize SIFT &\scriptsize 45.01$\pm$0.76 &\scriptsize 55.39$\pm$0.42 &\scriptsize 60.51$\pm$0.60 &\scriptsize 64.25$\pm$0.72 &\scriptsize 66.29$\pm$0.71 &\scriptsize 68.17$\pm$0.98 \\
\hline \scriptsize HSV-SIFT &\scriptsize 33.96$\pm$0.96 &\scriptsize 44.06$\pm$0.40 &\scriptsize 50.48$\pm$0.60 &\scriptsize 54.42$\pm$0.63 &\scriptsize 57.76$\pm$0.94 &\scriptsize 59.47$\pm$1.31 \\
\hline \scriptsize YCbCr-SIFT &\scriptsize \textbf{46.48 $\pm$ 0.91} &\scriptsize \textbf{56.97}$\pm$\textbf{0.60} &\scriptsize \textbf{62.09}$\pm$ \textbf{0.31} &\scriptsize \textbf{65.45}$\pm$\textbf{0.63} &\scriptsize \textbf{68.17}$\pm$\textbf{0.76} &\scriptsize \textbf{69.18}$\pm$\textbf{1.19} \\
\hline \scriptsize Opponent-SIFT &\scriptsize 27.00$\pm$0.48 &\scriptsize 35.07$\pm$0.58 &\scriptsize 39.31$\pm$0.55 &\scriptsize 41.93$\pm$0.99 &\scriptsize 44.21$\pm$1.06 &\scriptsize 45.87$\pm$0.74 \\
\hline \scriptsize rg-SIFT &\scriptsize 32.51$\pm$0.56 &\scriptsize 41.70$\pm$0.88 &\scriptsize 46.82$\pm$0.48 &\scriptsize 50.35$\pm$0.40 &\scriptsize 53.15$\pm$0.83 &\scriptsize 55.18$\pm$1.09 \\
\hline \scriptsize C-SIFT &\scriptsize 32.67$\pm$0.52 &\scriptsize 41.90$\pm$0.43 &\scriptsize 47.87$\pm$0.56 &\scriptsize 51.02$\pm$0.59 &\scriptsize 54.05$\pm$0.69 &\scriptsize 55.72$\pm$0.88 \\
\hline
\end{tabular}
\end{table*}
\subsection{Assessment of Color Descriptors on the Caltech-256 Dataset} \label{sec:Caltech256}
A more complex dataset, Caltech-256 \cite{griffin2007caltech}, is also employed for the experiments. It consists of 256 object classes and totaly 30,607 images, which have much higher intra-class variability and object location variability compared with the images in Caltech-101. Similar to section \ref{sec:Caltech101}, the gray images are also removed for fair comparison of various CSIFT/SIFT descriptors. Since there are at least 80 color images per category, no more image is added.

\begin{figure}
\centering
  \includegraphics[width=3in]{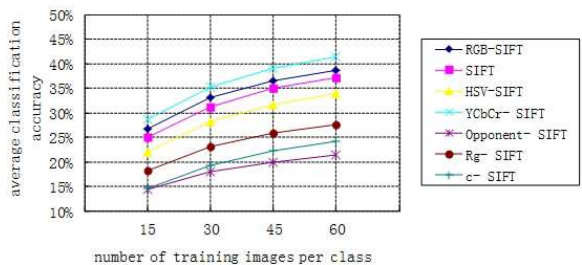}\\
  \caption{The different number of training images per class on the classification performance.}\label{fig:figure2}
\end{figure}

In each experiment, we randomly select $n$ ($n \in\{15, 30, 45, 60\}$ is fixed for each experiment) images from every category for training and leaving the remainders for testing. For the sake of simplicity, the codebook size is fixed to 4096 (according to our experience, it produces the best classification performance). The images were resized to keep the maximum size of height and width no larger than 300 pixels with conserved aspect ratio. The details of classification results are show in Table \ref{tab:table2} and Figure \ref{fig:figure2}. Among all these descriptors, YCbCr-SIFT produces the best performance as well. In case 60 random selected training images of each category are used, YCbCr-SIFT achieve the average classification accuracy of $41.3\%$; moreover, RGB-SIFT also provides the second best average classification accuracy ($38.7\%$). Compared with the performance of luminance-based SIFT descriptors, CSIFT brought approximately $4\%$ enhancement in regard of average classification accuracy, which can be significant in many image classification tasks.

\begin{table}
\centering
\caption{Classification rate$(\%)$ comparison on Caltech-256}
\label{tab:table2}
\begin{tabular}{|p{1.1cm}| p{1.1cm}| p{1.1cm}| p{1.1cm}| c|}
\hline \scriptsize Training images & \scriptsize \centering{15} & \scriptsize \centering{30} &\scriptsize \centering 45 & \scriptsize 60\\
\hline \scriptsize RGB-SIFT &\scriptsize 26.70$\pm$0.33 &\scriptsize33.04$\pm$0.22&\scriptsize 36.56$\pm$0.32&\scriptsize38.71$\pm$0.38 \\
\hline \scriptsize SIFT &\scriptsize 25.06$\pm$0.07 &\scriptsize 31.22$\pm$0.24 &\scriptsize 34.92$\pm$0.39 &\scriptsize 37.22$\pm$0.35  \\
\hline \scriptsize HSV-SIFT &\scriptsize 21.95$\pm$0.30 &\scriptsize 28.18$\pm$0.22 &\scriptsize 31.79$\pm$0.28 &\scriptsize 34.03$\pm$0.29 \\
\hline \scriptsize YCbCr-SIFT &\scriptsize \textbf{28.58$\pm$0.32} &\scriptsize \textbf{35.20$\pm$0.18} &\scriptsize \textbf{38.97$\pm$0.34} &\scriptsize \textbf{41.31$\pm$0.27} \\
\hline \scriptsize Opponent-SIFT &\scriptsize 14.37$\pm$0.24 &\scriptsize 17.92$\pm$0.22 &\scriptsize 20.0$\pm$0.20 &\scriptsize 21.43$\pm$0.45 \\
\hline \scriptsize Rg- SIFT &\scriptsize 18.16$\pm$0.24 &\scriptsize 22.98$\pm$0.26 &\scriptsize 25.88$\pm$0.36 &\scriptsize 27.63$\pm$0.31 \\
\hline \scriptsize c- SIFT &\scriptsize 14.56$\pm$0.18 &\scriptsize 19.30$\pm$0.22 &\scriptsize 22.13$\pm$0.19 &\scriptsize 24.19$\pm$0.27 \\
\hline
\end{tabular}
\end{table}

\section{Further Evaluations} \label{sec:Further}
The experimental results of section \ref{sec:Caltech101} and \ref{sec:Caltech256} show that, among the different CSIFT descriptors, YCbCr-SIFT and RGB-SIFT achieve better image classification performance than the state-of-the-art luminance-based SIFT. While, it is well-known that choosing different codebooks size, different numbers of neighbors in LLC and different pooling methods will affect the final classification results. In this section, further evaluations are carried out for more comprehensive studies of these two CSIFT descriptors.
\subsection{Impact of Codebook Size} \label{sec:Codebook_size}
\begin{figure}
  \includegraphics[width=3.1in]{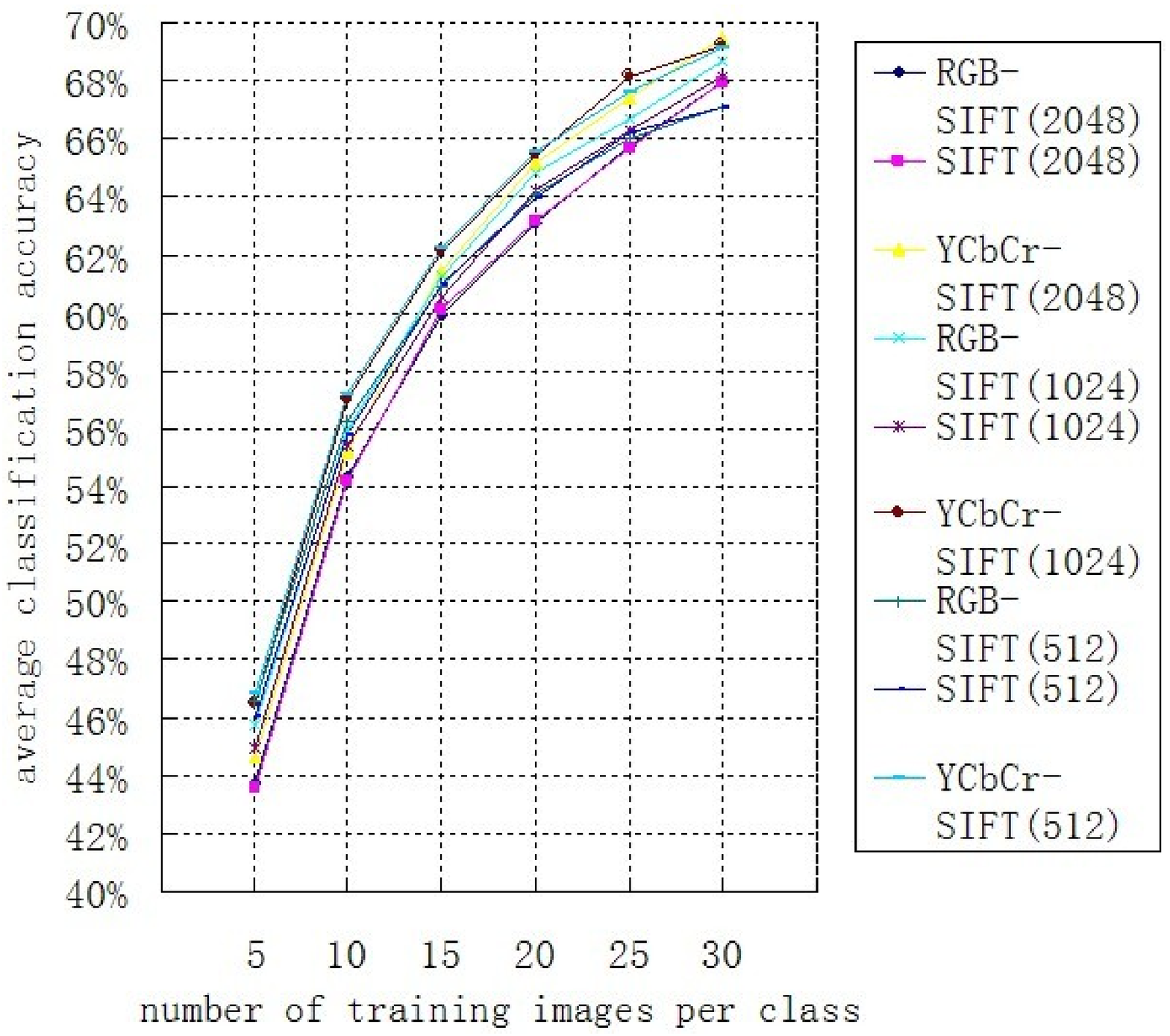}\\
  \caption{The different number of training images per class on the classification performance.}\label{fig:figure3}
\end{figure}

Firstly, we test the impacts of different codebook sizes (512, 1024 and 2048) using the Caltech-101 dataset. As discussed in section \ref{sec:Exp}, the codebooks are trained by the K-Means clustering algorithm. Different numbers (5, 10, $\dots$, 30) of training images per category are evaluated. The number of neighbors in LLC is set as 5. The corresponding results are presented in Table \ref{tab:table3}, Table \ref{tab:table4} , Table \ref{tab:table5} and Figure \ref{fig:figure3}. YCbCr-SIFT descriptor outperforms the others in all the tests. In most cases, the highest classification accuracy is obtained by using coodbook of size 1024. However, when the codebook of size 2048 is utilized, the classification accuracies decrease (except YCbCr-SIFT descriptor with 30 training images per category). It may be caused by the over-completeness of the codebooks, which results in large deviations in representing similar local features. It is interesting to notice that, by using more training data, the problem of over-completeness might be overcome. For the instance, YCbCr-SIFT descriptor with codebooks of size 2048 and 30 training images per category achieves the highest average classification accuracy.
\subsection{Impact of Different Number of Neighbors} \label{sec:neigbor}
\begin{figure}
  \includegraphics[width=3.1in]{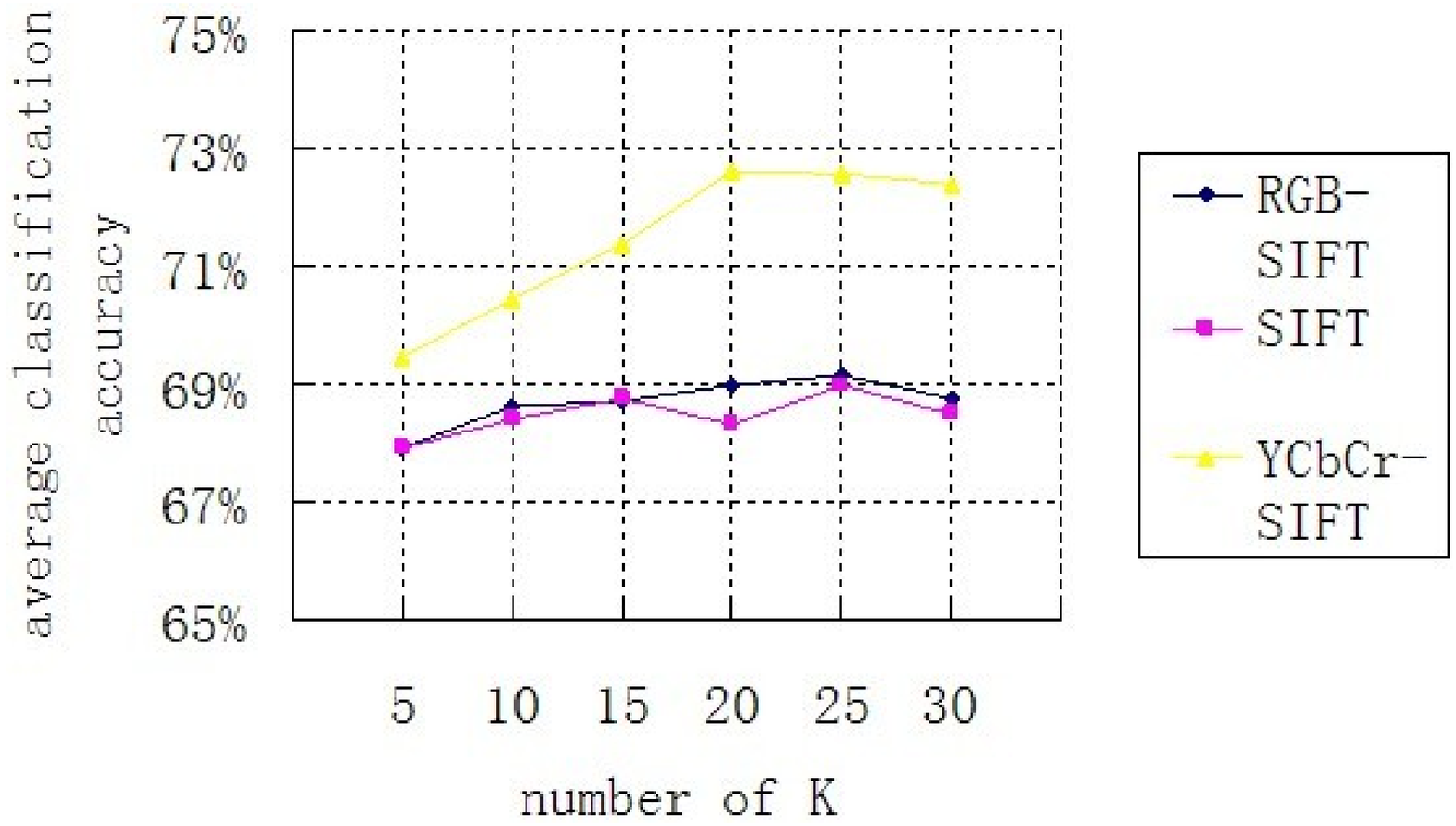}\\
  \caption{The different number of training images per class on the classification performance.}\label{fig:figure4}
\end{figure}

The performances of the proposed algorithm using different number of neighbors $K$ in LLC are also estimated. The codebook size is fixed as 1024, the number of training image per category is 30. The results are shown in Table \ref{tab:table1} and Figure \ref{fig:figure4}. With the increase of the neighbor number $K$ in LLC, the classification accuracy takes on the trend of rising first, then drops after $K \geq 25$. The highest average classification accuracy is obtained by using YCbCr-SIFT descriptor($72.59\%$). In contrast to the highest classification result of SIFT ($69.18\%$), more than $3\%$ improvement is achieved.
\subsection{Comparison of Pooling Methods}
\begin{figure}
  \includegraphics[width=3.1in]{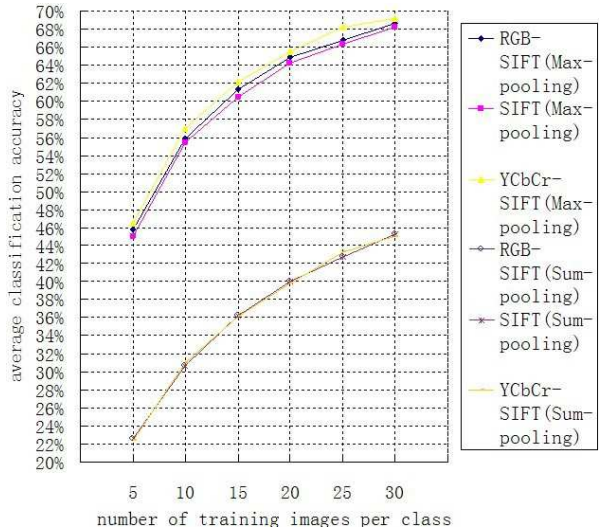}\\
  \caption{Impact of different pooling methods.}\label{fig:pooling}
\end{figure}

Besides the max-pooling method, sum-pooling is another choice which can also be used to summarize the features of each SPM layer. Table \ref{tab:table7}, Table \ref{tab:table8} show the experimental results using the two methods respectively. In Figure \ref{fig:pooling} they are illustrated together for comparison. The codebook size is 1024. The number of neighbors used in LLC is 5. It can be noticed that the max-pooling method significantly outperforms sum-pooling.
\begin{equation}\label{eq:max-pooling}
    Max: v_{j}=\max(v_{1},v_{2},\ldots,v_{i})
\end{equation}
\begin{equation}\label{eq:sum-pooling}
    Sum: v_{j}=v_{1}+v_{2}+\ldots+v_{i}
\end{equation}

As can be seen from Figure.\ref{fig:pooling}, the best performance is achieved by the combination of ``\emph{max-pooling}'' and ``\emph{$\ell^{2}$ normalization}''.

\begin{table*}
\centering
\caption{The codebooks of size 512}
\label{tab:table3}
\begin{tabular}{|l|c|c|c|c|c|c|} 
\hline \scriptsize Training images & \scriptsize 5 & \scriptsize 10 &\scriptsize 15 &\scriptsize 20 &\scriptsize 25 &\scriptsize 30 \\
\hline \scriptsize SIFT &\scriptsize 46.01$\pm$0.65 &\scriptsize 55.81$\pm$0.41 &\scriptsize 60.98$\pm$0.50 &\scriptsize 63.99$\pm$0.97 &\scriptsize 66.23$\pm$0.49 &\scriptsize 67.10$\pm$1.10 \\
\hline \scriptsize RGB-SIFT &\scriptsize 46.57$\pm$0.59 &\scriptsize 56.28$\pm$0.60 &\scriptsize 60.92$\pm$0.45 &\scriptsize 64.10$\pm$0.62 &\scriptsize 66.01$\pm$0.82 &\scriptsize 67.10$\pm$1.26 \\
\hline \scriptsize YCbCr-SIFT &\scriptsize \textbf{46.81}$\pm$\textbf{0.81} &\scriptsize \textbf{57.18}$\pm$\textbf{0.39} &\scriptsize \textbf{62.25}$\pm$\textbf{0.56} &\scriptsize \textbf{65.53}$\pm$\textbf{0.65} &\scriptsize \textbf{67.62}$\pm$\textbf{0.61} &\scriptsize \textbf{69.16}$\pm$\textbf{0.80} \\
\hline
\end{tabular}
\end{table*}

\begin{table*}
\centering
\caption{The codebooks of size 1024}
\label{tab:table4}
\begin{tabular}{|l|c|c|c|c|c|c|} 
\hline \scriptsize Training images & \scriptsize 5 & \scriptsize 10 &\scriptsize 15 &\scriptsize 20 &\scriptsize 25 &\scriptsize 30 \\
\hline \scriptsize SIFT &\scriptsize 45.01$\pm$0.76 &\scriptsize 55.39$\pm$0.42 &\scriptsize 60.51$\pm$0.60 &\scriptsize 64.25$\pm$0.72 &\scriptsize 66.29$\pm$0.71 &\scriptsize 68.17$\pm$0.98 \\
\hline \scriptsize RGB-SIFT &\scriptsize 45.77$\pm$1.02 &\scriptsize 55.90$\pm$0.69 &\scriptsize 61.26$\pm$0.84 &\scriptsize 64.84$\pm$0.68 &\scriptsize 66.70$\pm$0.81 &\scriptsize 68.65$\pm$1.13 \\
\hline \scriptsize YCbCr-SIFT &\scriptsize \textbf{46.48 $\pm$ 0.91} &\scriptsize \textbf{56.97}$\pm$\textbf{0.60} &\scriptsize \textbf{62.09}$\pm$ \textbf{0.31} &\scriptsize \textbf{65.45}$\pm$\textbf{0.63} &\scriptsize \textbf{68.17}$\pm$\textbf{0.76} &\scriptsize \textbf{69.18}$\pm$\textbf{1.19} \\
\hline
\end{tabular}
\end{table*}

\begin{table*}
\centering
\caption{The codebooks of size 2048}
\label{tab:table5}
\begin{tabular}{|l|c|c|c|c|c|c|}
\hline \scriptsize Training images & \scriptsize 5 & \scriptsize 10 &\scriptsize 15 &\scriptsize 20 &\scriptsize 25 &\scriptsize 30 \\
\hline \scriptsize SIFT &\scriptsize 43.56$\pm$0.78 &\scriptsize 54.18$\pm$0.78 &\scriptsize 60.08$\pm$0.72 &\scriptsize 63.18$\pm$0.54 &\scriptsize 65.68$\pm$0.63 &\scriptsize 67.91$\pm$1.21\\
\hline \scriptsize RGB-SIFT &\scriptsize 43.79$\pm$0.91 &\scriptsize 54.33$\pm$0.55 &\scriptsize 59.89$\pm$0.73 &\scriptsize 63.07$\pm$0.94 &\scriptsize 65.77$\pm$0.73 &\scriptsize 67.94$\pm$0.79 \\
\hline \scriptsize YCbCr-SIFT &\scriptsize \textbf{44.62}$\pm$\textbf{0.75} &\scriptsize \textbf{55.21}$\pm$\textbf{0.51} &\scriptsize \textbf{61.42}$\pm$\textbf{0.33} &\scriptsize \textbf{65.13}$\pm$\textbf{0.66} &\scriptsize \textbf{67.42}$\pm$\textbf{0.64} &\scriptsize \textbf{69.45}$\pm$\textbf{0.84} \\
\hline
\end{tabular}
\end{table*}

\begin{table*}
\centering
\caption{ Comparison on the sizes of the neighborhood size}
\label{tab:table6}
\begin{tabular}{|l|c|c|c|c|c|c|}
\hline \scriptsize Number of K & \scriptsize 5 & \scriptsize 10 &\scriptsize 15 &\scriptsize 20 &\scriptsize 25 &\scriptsize 30 \\
\hline \scriptsize SIFT &\scriptsize 67.91$\pm$1.21 &\scriptsize 68.41$\pm$1.03 &\scriptsize 68.74$\pm$0.94 &\scriptsize 68.31$\pm$0.84 &\scriptsize 68.99$\pm$0.86 &\scriptsize 68.51$\pm$1.17\\
\hline \scriptsize RGB-SIFT &\scriptsize 67.94$\pm$0.79 &\scriptsize 68.61$\pm$0.82 &\scriptsize 68.72$\pm$0.89 &\scriptsize 68.99$\pm$0.71 &\scriptsize 69.18$\pm$1.1 &\scriptsize 68.78$\pm$0.13 \\
\hline \scriptsize YCbCr-SIFT &\scriptsize \textbf{69.45}$\pm$\textbf{0.84} &\scriptsize \textbf{70.44}$\pm$\textbf{1.03} &\scriptsize \textbf{71.37}$\pm$\textbf{0.72} &\scriptsize \textbf{72.59}$\pm$\textbf{0.63} &\scriptsize \textbf{72.56}$\pm$\textbf{1.22} &\scriptsize \textbf{72.39}$\pm$\textbf{1.47} \\
\hline
\end{tabular}
\end{table*}


\begin{table*}
\centering
\caption{The performance of $max-pooling$}
\label{tab:table7}
\begin{tabular}{|l|c|c|c|c|c|c|}
\hline \scriptsize Training images & \scriptsize 5 & \scriptsize 10 &\scriptsize 15 &\scriptsize 20 &\scriptsize 25 &\scriptsize 30 \\
\hline \scriptsize SIFT &\scriptsize 45.01$\pm$0.76 &\scriptsize 55.39$\pm$0.42 &\scriptsize 60.51$\pm$0.60 &\scriptsize 64.25$\pm$0.72 &\scriptsize 66.29$\pm$0.71 &\scriptsize 68.17$\pm$0.98 \\
\hline \scriptsize RGB-SIFT &\scriptsize 45.77$\pm$1.02 &\scriptsize 55.90$\pm$0.69 &\scriptsize 61.26$\pm$0.84 &\scriptsize 64.84$\pm$0.68 &\scriptsize 66.70$\pm$0.81 &\scriptsize 68.65$\pm$1.13 \\
\hline \scriptsize YCbCr-SIFT &\scriptsize \textbf{46.48 $\pm$ 0.91} &\scriptsize \textbf{56.97}$\pm$\textbf{0.60} &\scriptsize \textbf{62.09}$\pm$ \textbf{0.31} &\scriptsize \textbf{65.45}$\pm$\textbf{0.63} &\scriptsize \textbf{68.17}$\pm$\textbf{0.76} &\scriptsize \textbf{69.18}$\pm$\textbf{1.19} \\
\hline
\end{tabular}
\end{table*}

\begin{table*}
\centering
\caption{The performance of $sum-pooling$}
\label{tab:table8}
\begin{tabular}{|l|c|c|c|c|c|c|}
\hline \scriptsize Training images & \scriptsize 5 & \scriptsize 10 &\scriptsize 15 &\scriptsize 20 &\scriptsize 25 &\scriptsize 30 \\
\hline \scriptsize SIFT &\scriptsize 22.14$\pm$0.78 &\scriptsize 30.14$\pm$0.85 &\scriptsize \textbf{36.38$\pm$0.47} &\scriptsize 38.98$\pm$1.03 &\scriptsize 41.86$\pm$0.61 &\scriptsize 45.0$\pm$1.06 \\
\hline \scriptsize RGB-SIFT &\scriptsize \textbf{22.67}$\pm$\textbf{0.73} &\scriptsize 30.64$\pm$0.63 &\scriptsize 36.26$\pm$0.87 &\scriptsize \textbf{40.04$\pm$0.41} &\scriptsize 42.71$\pm$0.82 &\scriptsize \textbf{45.24$\pm$0.77} \\
\hline \scriptsize YCbCr-SIFT &\scriptsize 22.42$\pm$1.06 &\scriptsize \textbf{31.04}$\pm$\textbf{0.65} &\scriptsize 36.12$\pm$0.62 &\scriptsize 39.83$\pm$0.83 &\scriptsize \textbf{43.28$\pm$0.87} &\scriptsize 45.10$\pm$1.33 \\
\hline
\end{tabular}
\end{table*}

\section{Conclusion} \label{sec:conclusion}
In this article, CSIFT descriptors are introduced to improve the state-of-the-art \emph{Locality-constrained Linear Coding} (LLC) based image classification system. Different kinds of CSIFT descriptors are implemented and evaluated with varies settings of the parameters. Real experiments have demonstrated     that, by utilizing color information, considerable improvements can be obtained. Among the CSIFT descriptors, YCrCb-SIFT descriptor achieves the most stable and accurate image classification performance. Compared with the highest average classification accuracy achieved by using luminance-base SIFT descriptors, YCrCb-SIFT descriptor acquired approximate $3\%$ increase on the Caltech-101 dataset (see section \ref{sec:neigbor}) and approximate $4\%$ increase on the Caltech-256 dataset (see section \ref{sec:Caltech256}). Besides the YCrCb-SIFT descriptor, RGB-SIFT descriptor also provides favorable performance. As one of the most representative SR based image classification algorithms, the improvements achieve on LLC show that using CSIFT descriptors is an approach with good potential to enhance state-of-the-art SR based image classification systems. On the other hand, although be reported can achieve invariant or discriminatory object recognition, we found that the performances of some others CSIFT descriptors are not as good as expected. One potential solution is combing different CSIFT descriptor to build a better one, we will try to study it in the future work.

\section*{Acknowledgments}
This work is supported by the National Natural Science Foundation of China (No.61003143) and the Fundamental Research Funds for Central Universities (No.SWJTU12CX094).


\end{document}